\documentclass[runningheads]{llncs}
\usepackage[T1]{fontenc}
\usepackage{graphicx,verbatim}
\usepackage{amsmath}
\usepackage{amsfonts}
\usepackage{booktabs}

\usepackage{microtype}

\usepackage{marvosym}

\begin{document}
\title{Self-Supervised Temporal Regularization for Landmark-Based Cardiac Segmentation with Automatic AHA Regional Mapping}
\titlerunning{Self-Supervised Temporal Regularization with Automatic AHA Mapping}
\author{David Montalvo-García\inst{1,2}$^{(\textrm{\Letter})}$ \and
Nicolás Gaggion\inst{3,4,5} \and
María J. Ledesma-Carbayo\inst{1,2} \and
Enzo Ferrante\inst{5}$^{(\textrm{\Letter})}$}
% index{Montalvo-García, David}
% index{Gaggion, Nicolás}
% index{Ledesma-Carbayo, María J.}
% index{Ferrante, Enzo}
%
\authorrunning{D. Montalvo-García et al.}

\institute{Biomedical Image Technologies, ETSI Telecomunicación, Universidad\\ Politécnica de Madrid, Madrid, Spain\\
\email{david.montalvo@upm.es} \and
Centro de Investigación Biomédica en Red de Bioingeniería, Biomateriales y Nanomedicina (CIBER-BBN), Instituto de Salud Carlos III, Madrid, Spain \and
APOLO Biotech, Ciudad Autónoma de Buenos Aires, Argentina \and
Departamento de Computación, Universidad de Buenos Aires, Ciudad Autónoma de Buenos Aires, Argentina \and
Instituto de Ciencias de la Computación (ICC), CONICET-Universidad de Buenos Aires, Ciudad Autónoma de Buenos Aires, Argentina\\
\email{eferrante@dc.uba.ar}}
  
\maketitle              % typeset the header of the contribution

\vspace{-3mm}

\begin{abstract}
Graph-based cardiac segmentation with implicit anatomical correspondences provides topological guarantees and population-level analysis capabilities, but models trained on independent frames of image sequences exhibit temporal discontinuities that affect reliable clinical measurements, particularly in cardiac ultrasound. In this work, we introduce self-supervised temporal regularization as a post-training refinement stage that exploits the temporal coherence in image sequences to enforce consistent cardiac segmentation and motion estimation over time, without requiring per-frame annotations. By penalizing velocity and acceleration discontinuities across consecutive frames, our method achieves temporally consistent segmentations while maintaining the learned anatomical correspondences. We further leverage these correspondences to automatically map landmarks to the AHA 17-segment clinical standard, enabling standardized regional assessment and detection of pathological myocardial motion patterns. Validation on CAMUS dataset demonstrates the clinical utility of combining temporal consistency with automatic regional mapping. The code is publicly available at
\mbox{\url{https://github.com/david-montalvoo/MaskHybridGNet-TempReg}}.

% Authors must provide keywords and are not allowed to remove this Keyword section.

\end{abstract}

\section{Introduction}
\vspace{-1mm}

Deep learning has transformed medical image segmentation, with U-Net architectures~\cite{ronneberger2015u} and Vision Transformers~\cite{valanarasu2021medical} achieving remarkable performance across modalities~\cite{litjens2017survey}. However, pixel-based methods trained with cross-entropy or Dice losses~\cite{milletari2016v} lack explicit anatomical knowledge, producing segmentations with topological inconsistencies, holes, and irregular boundaries under challenging conditions~\cite{choudhary2020advancing,cohen2020limits}. Graph-based segmentation addresses these limitations by representing anatomical boundaries as connected graphs with guaranteed topology~\cite{boussaid2014discriminative}. Recent advances in implicit anatomical correspondence learning~\cite{gaggion2025contour} enable training on standard pixel-wise masks using Chamfer distance, where the $i$-th predicted landmark emerges to consistently represent the same anatomical location across patients, providing competitive accuracy with superior robustness and enabling automatic atlas generation.

Despite anatomical consistency across patients, these models process frames independently, ignoring temporal coherence. For cardiac imaging, this produces frame-to-frame discontinuities that degrade ejection fraction estimation, prevent accurate wall motion tracking required for clinical assessment, introduce artificial jitter incompatible with physiological cardiac dynamics, and limit translation to clinical workflows. Existing temporal consistency approaches rely on post-processing with optical flow~\cite{ledesma2005spatio}, Kalman filtering, or autoencoder refinement~\cite{painchaud2022echocardiography}. While effective, these operate independently of the segmentation model and cannot leverage learned anatomical structure.

In this work, we introduce self-supervised temporal regularization as a post-training refinement stage for landmark-based cardiac segmentation. Starting from a model with established implicit correspondences, we fine-tune on ultrasound sequences by penalizing velocity and acceleration discontinuities between consecutive frames. This exploits the fundamental principle that anatomically-corresponding landmarks should move smoothly due to physiological constraints. Critically, this requires no per-frame annotations: the temporal behavior itself provides the supervisory signal. We further leverage learned correspondences to automatically map landmarks to the AHA 17-segment standard~\cite{cerqueira2002standardized}, enabling clinically-interpretable regional analysis. This combination of temporal consistency and automatic regional mapping allows detection of pathological wall motion patterns such as regional dyssynchrony and hypokinesis, making the approach directly applicable to clinical workflows.
\vspace{-1mm}

\subsection{Related Work}
\vspace{-1mm}

\noindent \textbf{Landmark-Based Cardiac Segmentation.} Point Distribution Models~\cite{cootes1992training} and Active Shape Models pioneered landmark-based anatomical modeling through statistical shape variation. HybridGNet~\cite{gaggion_miccai,gaggion} combined CNNs with spectral graph convolutions for variational cardiac segmentation, demonstrating superior domain shift robustness but requiring manually annotated landmarks with anatomical correspondences. Recent work~\cite{gaggion2025contour} eliminates this requirement by training on standard segmentation masks, where the $i$-th landmark emerges to consistently represent the same anatomical location through optimization with Chamfer distance and edge regularization. This enables topologically-guaranteed segmentation with automatic correspondence discovery. However, temporal consistency remains unaddressed as models process frames independently.\\

\noindent \textbf{Temporal Consistency in biomedical image sequences.} Classical approaches employ registration or optical flow based techniques~\cite{ledesma2005spatio} as well as Kalman filtering for exploiting temporal coherence. Ledesma-Carbayo et al.~\cite{ledesma2005spatio} introduced  a B-spline based spatio-temporal registration framework to exploit temporal coherence for the estimation of cardiac motion on ultrasound sequences. Sundar et al.~\cite{sundar20094d} proposed simultaneous 4D registration where all cardiac phases are jointly optimized for temporal smoothness. Recent deep learning methods include recurrent architectures and test-time refinement. Painchaud et al.~\cite{painchaud2022echocardiography} use a constrained autoencoder for post-hoc correction of temporal inconsistencies in the attributes driven latent representation of echocardiography segmentations, operating on outputs from any segmentation method. Our work differs by integrating temporal regularization as a fine-tuning stage for landmark-based segmentation that leverages learned anatomical correspondences, enabling the model to refine its temporal representations while preserving spatial accuracy and structure correspondence.

\section{Methods}

\subsection{Background: Implicit Correspondence Learning} 
\vspace{-1mm}

Our framework builds on the implicit correspondence learning approach~\cite{gaggion2025contour} where a graph-based VAE model learns to map input images $\mathbf{I}$ to fixed-size landmark graphs $\mathbf{G} = \langle V, \mathbf{A}, \mathbf{X} \rangle$. Here $V$ represents a set of $M$ nodes corresponding to anatomical landmarks, $\mathbf{A}$ is an adjacency matrix encoding connectivity between organ contours, and $\mathbf{X} \in \mathbb{R}^{M \times 2}$ contains the landmark coordinates. The encoder maps images to a latent distribution $\mathbf{z} \sim \mathcal{N}(\boldsymbol{\mu}, \boldsymbol{\sigma}^2)$, which the graph-convolutional decoder transforms into landmark coordinates. Completing the Mask-HybridGNet Dual architecture, an auxiliary convolutional decoder generates dense pixel-level masks and routes its multi-scale features directly to the graph decoder via image-to-graph skip connections.

The graph adjacency matrix is constructed automatically from an atlas derived from CAMUS ground-truth segmentation masks to represent the left ventricular endocardium, epicardium, and left atrium as a unified graph. This structure enforces anatomical constraints through shared boundaries: the left ventricular endocardium shares nodes with the inner myocardial wall, and the basal endocardium shares nodes with the left atrium at the mitral valve plane. 

We train the base model with a compound loss functions which uses Chamfer distance ($\mathcal{L}_{CD}$) to match variable-length ground truth contours $\mathbf{P}$ with the fixed-size predictions $\mathbf{G}$, combined with edge regularization for uniform spacing between landmarks ($\mathcal{L}_{uniform}$), elasticity regularization to encourage compact contours ($\mathcal{L}_{elastic}$), and curvature regularization ($\mathcal{L}_{smooth}$) to promote local smoothness by penalizing abrupt direction changes between consecutive edges, along with a standard VAE Kullback–Leibler divergence on the variational latents ($\mathcal{L}_{KL}$). We also include a Dice pixel-loss ($\mathcal{L}_{pixel}$) computed by comparing the ground-truth masks with both, the rasterized graph-contour predictions and the dense
output of the auxiliary CNN decoder. This process discovers anatomical correspondences where landmark $i$ represents consistent anatomy across all subjects. The standard frame-independent training objective is:
\begin{equation}
\mathcal{L}_{base} = \mathcal{L}_{CD} + \lambda_{KL}\mathcal{L}_{KL} + \lambda_e\mathcal{L}_{elastic} + \lambda_u\mathcal{L}_{uniform} + \lambda_s\mathcal{L}_{smooth} + \lambda_{p}\mathcal{L}_{pixel}  
\label{eq:loss_base}
\end{equation}

For a more detailed description of these loss terms and the Mask-HybridGNet Dual architecture we refer the reader to \cite{gaggion2025contour}.
After training with this objective, the model produces anatomically consistent segmentations with implicit correspondences. However, the combination of fixed-size predictions and Chamfer distance matching can cause temporal instability: landmarks may shift by one or two positions along the circular contour between consecutive frames, resulting in correspondence oscillations and abrupt temporal transitions.

\subsection{Self-Supervised Temporal Regularization}
\label{sec:selfsupervised_temporal_regularization}
\vspace{-1mm}

To address the temporal instability of the predicted landmarks while preserving learned correspondences, we introduce temporal regularization as a post-training refinement stage. Starting from a converged model with established anatomical correspondences, we fine-tune on video sequences by adding temporal constraints that exploit the principle that anatomically-corresponding landmarks should exhibit smooth motion profiles characteristic of physiological cardiac dynamics.

Let $\{\mathbf{I}^{(t)}\}_{t=1}^T$ denote a temporal sequence of $T$ consecutive frames from the same cardiac cycle. For each frame, the encoder produces latent code $\mathbf{z}_t \sim \mathcal{N}(\boldsymbol{\mu}_t, \boldsymbol{\sigma}_t^2)$ and the decoder generates a set of landmarks $\mathbf{X}(t)=\big\{x_i(t)\big\}_{i=1}^M\in \mathbb{R}^{M \times 2}$. We introduce two complementary temporal regularization terms that penalize motion discontinuities while maintaining the spatial accuracy learned during initial training.\\

 \noindent \textbf{Velocity regularization} encourages smooth trajectories for anatomically corresponding landmarks by penalizing large frame-to-frame displacements:

\begin{equation}
\mathcal{L}_{vel} = \frac{1}{M}\sum_{i=1}^{M} \Bigg(\frac{1}{T-1}\sum_{t=1}^{T-1} \big\Vert\,\mathbf{x}_i(t+1) - \mathbf{x}_i(t)\big\Vert^2\Bigg)
\label{eq:loss_vel}
\end{equation}

\noindent \textbf{Acceleration regularization} penalizes abrupt changes in velocity by minimizing the difference between consecutive velocity vectors, enforcing physiologically plausible acceleration and deceleration patterns:

\begin{equation}
\mathcal{L}_{accel} = \frac{1}{M}\sum_{i=1}^{M} \Bigg(\frac{1}{T-2}\sum_{t=1}^{T-2} \big\Vert\mathbf{x}_i(t+2) - 2\,\mathbf{x}_i(t+1) + \mathbf{x}_i(t)\big\Vert^2\Bigg)
\label{eq:loss_accel}
\end{equation}

\noindent \textbf{The refinement objective} combines the original spatial terms with temporal regularization:
\begin{equation}
\mathcal{L}_{refine} = \mathcal{L}_{base} + \lambda_v\mathcal{L}_{vel} + \lambda_a\mathcal{L}_{accel} 
\label{eq:loss_refine}
\end{equation}
where $\lambda_v, \lambda_a$ control the strength of temporal regularization.\\

\noindent \textbf{Training Procedure.} We implement an alternating optimization strategy to balance spatial accuracy with temporal consistency. Each training iteration processes a batch consisting of: (1) $N$ individual annotated frames optimized using $\mathcal{L}_{base}$, followed by (2) one temporal sequence of $T$ consecutive frames optimized using $\mathcal{L}_{vel}$ and $\mathcal{L}_{accel}$. This alternating scheme is essential to prevent the model from converging to trivial solutions that minimize temporal losses by predicting static contours with minimal motion. The sequence length $T$ is constrained by GPU memory requirements, as processing full sequences requires maintaining activations for all $T$ frames during backpropagation.

\vspace{-2mm}

\subsection{Automatic AHA Mapping}
\label{sec:aha_mapping}
\vspace{-1mm}

\begin{figure}[h]
\centering
\includegraphics[width=0.95\linewidth]{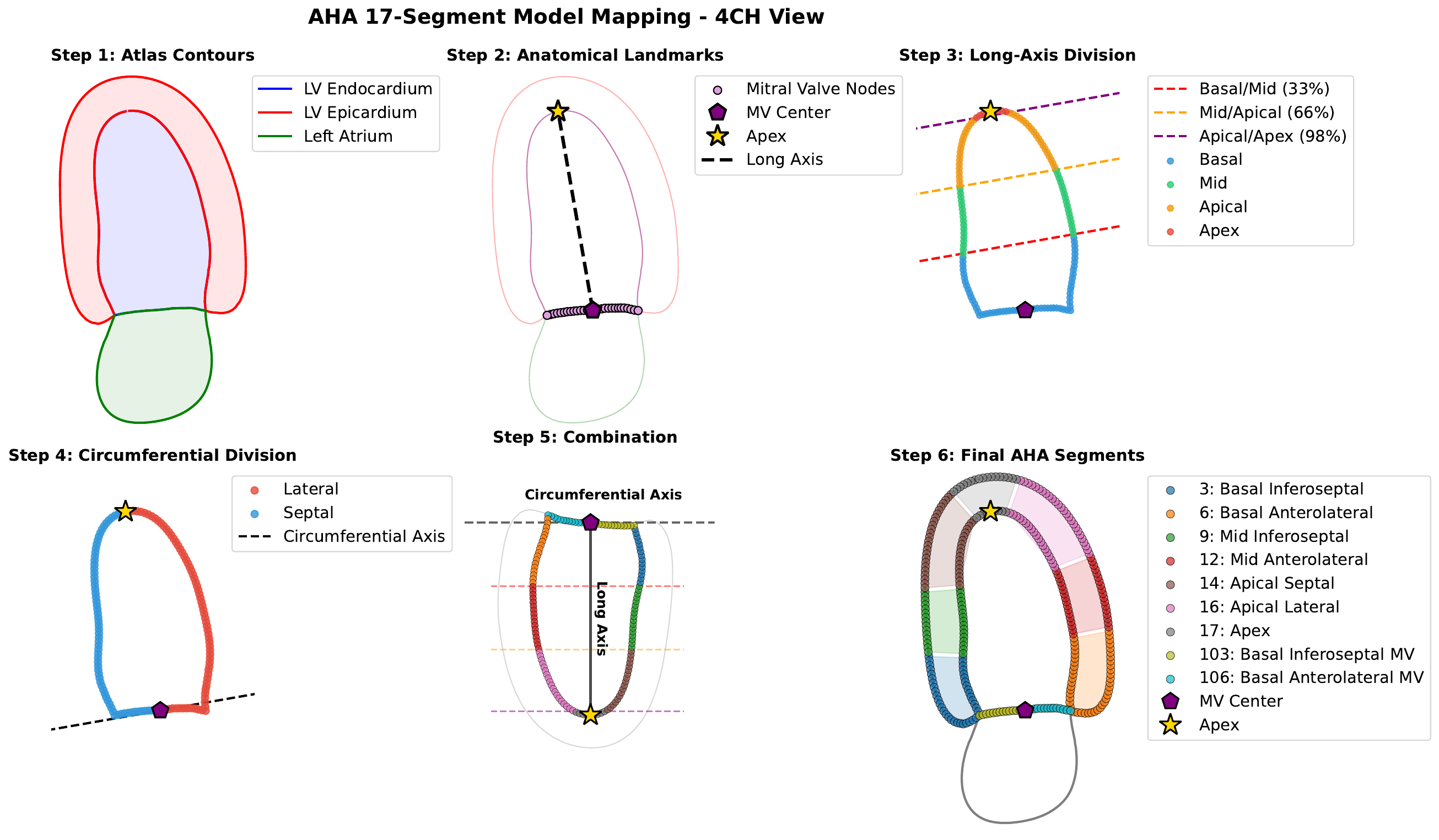}
\caption{\textbf{Automatic AHA mapping via implicit correspondence.} Step-by-step procedure for the apical 4-chamber view: (1) population atlas generation at ED; (2) anatomical landmark identification and long axis definition; (3) long-axis division into basal (0--33\%), mid (33--66\%), and apical (66--98\%) levels; (4) circumferential bisection; (5) combined classification; and (6) final AHA segment assignments. The 2-chamber view follows an identical procedure to map its respective anterior and inferior segments.}
\label{fig:aha_mapping}
\end{figure}

We leverage the learned implicit correspondences to automatically map landmarks to the standardized AHA 17-segment model \cite{cerqueira2002standardized}, translating established clinical echocardiography workflows \cite{lang2015recommendations} into our automated pipeline. Because the landmark $i$ consistently represents the same anatomical location across the dataset, this mapping is computed just once on a population-averaged reference atlas, constructed by averaging the spatial coordinates of all subjects at the end-diastolic frame, and applied universally. As illustrated for the 4-chamber view in Figure~\ref{fig:aha_mapping}, the procedure automatically identifies the mitral valve center from shared left ventricular and atrial nodes, establishing the primary long-axis vector toward the endocardial apex. The left ventricle is then divided along this long axis into basal (up to 33\%), mid-cavity (33\% to 66\%), and apical (66\% to 98\%) levels, reserving the final 2\% for the apex cap. These levels are perpendicularly bisected to separate the opposing walls (anterolateral and inferoseptal in the 4-chamber view; anterior and inferior in the 2-chamber view). Finally, epicardial nodes are assigned to the segment of their nearest endocardial neighbor.

\section{Experiments, results \& discussion}
\vspace{-2mm}

\noindent \textbf{Datasets and Evaluation}. The CAMUS dataset~\cite{leclerc2019camus} provides 500 patients with 2-chamber and 4-chamber apical echocardiography sequences of approximately 20 frames each. Expert annotations mark LV endocardium, epicardium, and left atrium at end-diastole and end-systole, with metadata including ejection fraction and image quality scores. We use the 400-50-50 train/val/test split provided. 

For spatial accuracy, we compute the Dice coefficient by rasterizing the predicted landmarks into binary masks, and the 95th percentile Hausdorff distance (HD95) to quantify boundary error while reducing sensitivity to outlier points. Clinical metrics were selected to reflect standard measures used in routine cardiac practice, and performance was evaluated by computing the mean absolute error (MAE) between the reference and predicted values for ejection fraction (EF), end-systolic volume (ESV), and end-diastolic volume (EDV), with ventricular volumes computed from masks using the biplane Simpson’s method.

To quantify the temporal regularity of the trajectories while excluding the expected displacements induced by cardiac motion, we use a jitter metric that captures high-frequency perturbations. Specifically, jitter is defined as the root-mean-square (RMS) deviation between each aligned trajectory and a low-pass filtered version of the same trajectory over time:

\begin{equation}
\text{Jitter} = \frac{1}{M} \sum_{i=1}^{M} \Bigg(\sqrt{\frac{1}{T}\sum_{t=1}^{T}\;\big\Vert \; \mathbf{x}_i(t) - \bar{\mathbf{x}}_i(t) \,\big\Vert ^2}\Bigg)
\end{equation}
where $\bar{\mathbf{x}}_i(t)$ denotes the smoothed trajectory obtained by applying a Savitzky--Golay filter independently to each aligned trajectory. As complementary measures of trajectory quality and smoothness, we also report frame-to-frame displacement (FTD), which quantifies the RMS excursion distance for each node
\begin{equation}
\text{FTD} = \frac{1}{M} \sum_{i=1}^{M} \Bigg(\sqrt{\frac{1}{T-1} \sum_{t=1}^{T-1} \big\Vert\, \mathbf{x}_i(t+1) - \mathbf{x}_i(t)\big\Vert^2}\Bigg)
\end{equation}
and RMS jerk, which measures the temporal variation of acceleration and penalizes abrupt motion changes using the 3rd-order finite difference

\begin{equation}
\text{Jerk} = \frac{1}{M} \sum_{i=1}^{M} \Bigg(\sqrt{\frac{1}{T-3} \sum_{t=1}^{T-3} \big\Vert\, \Delta^3\mathbf{x}_i(t) \big\Vert^2}\Bigg)
\end{equation}

\begin{figure}[h]
\centering
\includegraphics[width=\linewidth]{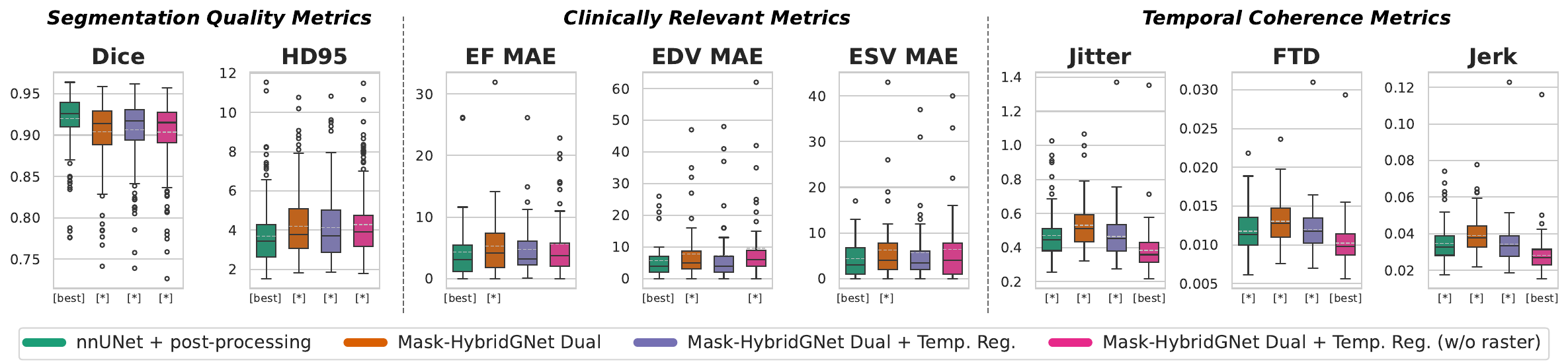}
\caption{\textbf{Quantitative comparison across segmentation, clinical, and temporal coherence metrics.} Boxplots report the distribution of Dice and HD95 (left), MAE for EF, EDV, and ESV (center), and Jitter, FTD and Jerk (right) computed from predicted landmark trajectories. The label [best] indicates the best mean performance for each metric (higher is better for Dice; lower is better for all others), and [*] denotes statistically significant differences with respect to the best-performing model for that metric according to the Wilcoxon signed-rank test}
\label{fig:boxplots}
\end{figure}

\noindent \textbf{Results.} We first trained Mask-HybridGNet Dual on CAMUS without temporal regularization, using the loss in Eq.~\ref{eq:loss_base} and ED and ES frames with ground-truth masks from both 2-chamber and 4-chamber views. We selected the dual branch architecture because it showed the best performance in the experiments reported in \cite{gaggion2025contour}. Figure~\ref{fig:boxplots} summarizes all spatial, clinical, and trajectory metrics. This baseline achieves strong segmentation performance (Dice and HD95), but relatively poor trajectory quality and smoothness metrics, consistent with the absence of temporal regularization during training.

For comparison, we also trained nnUNet~\cite{isensee2021nnunet} on the same CAMUS images and masks. Since nnUNet is a pixel-wise segmentation model and does not provide landmarks or temporal correspondences, trajectory-based temporal consistency metrics cannot be computed directly from its output. To enable this analysis, we trained a Mask-HybridGNet model that takes multiclass segmentation masks as input (instead of images) and predicts spatially consistent nodes using a unified multiclass contour. This enables the post-processing of nnUNet predictions over the cardiac cycle, converting them into trajectories and quantifying their temporal regularity. Although this pipeline enables trajectory analysis, its temporal coherence metrics show statistically significant differences compared to those of the proposed temporally regularized approach. (see Figure~\ref{fig:boxplots}).

Starting from the supervised Mask-HybridGNet Dual baseline, we then performed post-training temporal regularization following Section \ref{sec:selfsupervised_temporal_regularization}, using the loss in Eq.~\ref{eq:loss_refine}, which includes velocity (Eq.~\ref{eq:loss_vel}) and acceleration (Eq.~\ref{eq:loss_accel}) terms with weights selected by grid search on the validation set. We evaluated two variants, with and without rasterization in the $\mathcal{L}_{pixel}$ loss-term (Mask-HybridGNet Dual + Temp. Reg. and Mask-HybridGNet Dual + Temp. Reg. w/o Raster).

\begin{figure}[h]
\centering
\includegraphics[width=\linewidth]{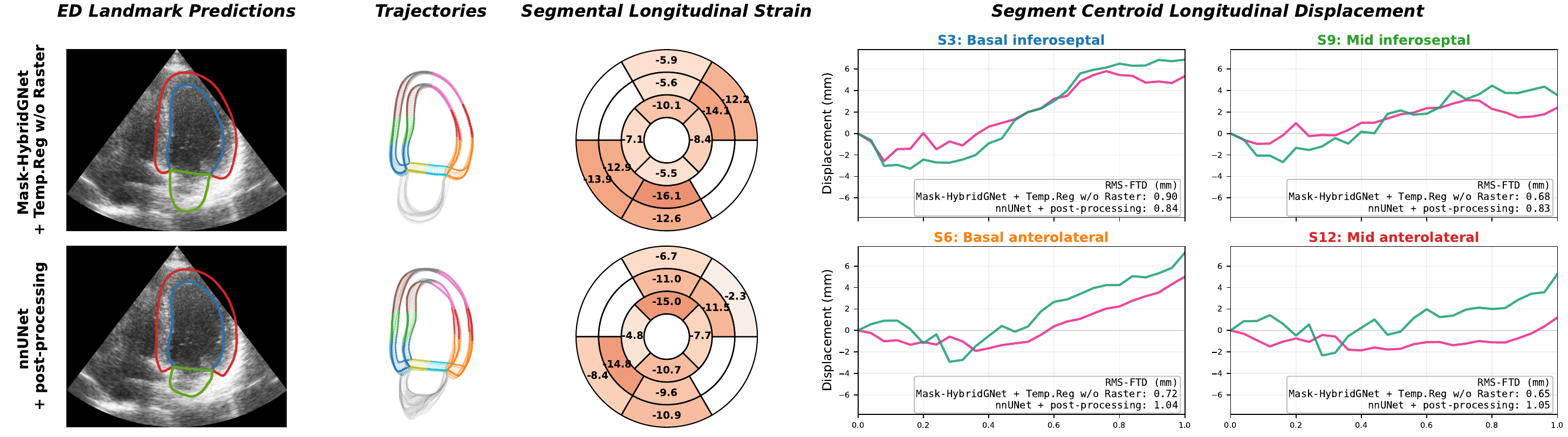}

\vspace{-3.2mm}
\noindent\rule{\linewidth}{0.2pt}
\vspace{-3.2mm}

\includegraphics[width=\linewidth]{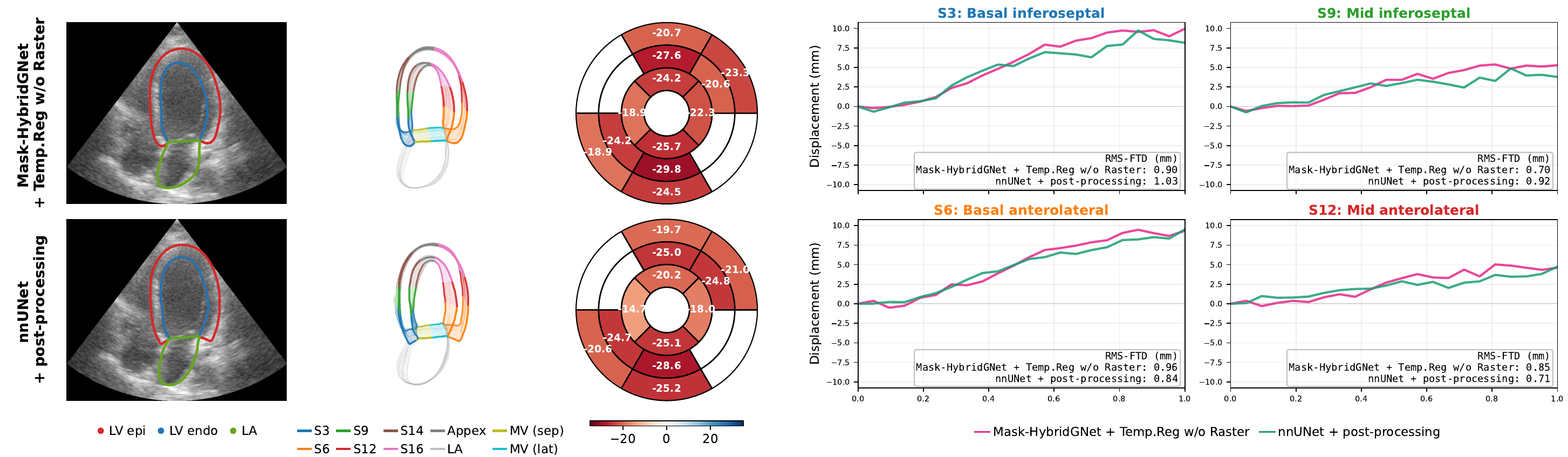}
\caption{\textbf{Qualitative comparison of regional motion trajectories and strain in two representative test cases.} For each patient, the figure compares \textit{Mask-HybridGNet + Temp.Reg. w/o Raster} and \textit{nnUNet + post-processing} using (from left to right) ED landmark predictions, full landmark trajectories (4CH) and segmental longitudinal strain based on AHA segments (4CH/2CH), and longitudinal displacement curves of selected AHA segment centroids (S3, S6, S9, and S12), with RMS-FTD values reported in each subplot. Top: Patient0051 (EF=29\%, reduced EF). Bottom: Patient0243 (EF=55\%, normal EF). The temporally regularized model yields smoother segment-centroid excursions and more regular trajectories, especially at the left atrium.}
\label{fig:trajectories}
\end{figure}

Among the evaluated models, Mask-HybridGNet Dual + Temp. Reg. w/o Raster achieves the best trajectory quality metrics, with smoother trajectories than both the non-regularized Mask-HybridGNet baseline and the nnUNet + post-processing pipeline. To perform a qualitative regional analysis, predicted landmarks were assigned to AHA segments using the self-supervised procedure described in Section~\ref{sec:aha_mapping}. Figure~\ref{fig:trajectories} shows AHA segment identification and landmark trajectories from ED to ES in two test cases, together with longitudinal displacements of the segment centroids. The temporally regularized model exhibits smoother segment-centroid trajectories than the post-processed nnUNet, consistent with the improvements observed in the temporal coherence metrics. As illustrated in Figure~\ref{fig:trajectories}, graph-based models with implicit correspondence learning show strong potential for downstream clinical tasks, including segment-wise AHA longitudinal strain estimation. The Supplementary Video shows the dynamic representation of the comparative performance for 6 test cases. 

\noindent \textbf{Discussion \& Conclusions.} In this work, we introduced a self-supervised post-training temporal regularization strategy for graph-based cardiac ultrasound segmentation that improves temporal coherence while preserving anatomical correspondences, allowing automatic AHA-based regional motion analysis. Our results show that high pixel-level segmentation accuracy does not necessarily ensure temporally consistent trajectories. Although nnUNet performs strongly on conventional spatial and clinical metrics, it lacks explicit landmark correspondences and temporal regularization, limiting direct trajectory-based analysis. Converting nnUNet masks into trajectories partially addresses this limitation, but yields inferior temporal quality compared to models enforcing consistency. The proposed regularized Mask-HybridGNet improves trajectory smoothness while maintaining spatial and clinical performance comparable to nnUNet, and naturally supports standardized regional motion analysis.\\

\noindent \textbf{Acknowledgments.} EF was supported by the Google Award for Inclusion Research, and a Googler Initiated Grant. MJL acknowledges the support of the Spanish Ministerio de Ciencia e Innovación, Agencia Estatal de Investigación, under grant PID2022-141493OB-I00 (10.13039/501100011033/MCIN/AEI/ERDF, UE), cofinanced by European Regional Development Fund (ERDF), ‘A way of making Europe, and partial funding from the MAGERIT-CM project (TEC-2024/COM-44), supported by the Research and Development Activities Program by the Comunidad de Madrid. DMG was supported by an FPU PhD fellowship funded by the Spanish Ministerio de Ciencia, Innovación y Universidades.\\

% \noindent \textbf{Disclosure of Interests.} The authors have no competing interests to declare.

\bibliographystyle{splncs04.bst}
\bibliography{Paper-5651}

\end{document}